\documentclass[final]{nesy2026}

\usepackage{times}
\usepackage{tikz}
\usetikzlibrary{arrows.meta, positioning, shadows, shadows.blur, calc}
\usepackage{multirow}
\usepackage{booktabs}
\usepackage{xcolor}
\usepackage{hyperref}
\usepackage{colortbl}
\usepackage{caption}
\usepackage{wrapfig}
\usepackage{footmisc}
\usepackage{pdfpages}

\makeatletter
\g@addto@macro\normalsize{%
  \setlength\abovedisplayskip{3pt plus 1pt minus 1pt}%
  \setlength\belowdisplayskip{3pt plus 1pt minus 1pt}%
  \setlength\abovedisplayshortskip{1pt plus 1pt}%
  \setlength\belowdisplayshortskip{1pt plus 1pt minus 1pt}%
}
\makeatother
\usepackage{titlesec}
\titlespacing*{\section}{0pt}{0.8ex plus .2ex minus .2ex}{0.4ex plus .1ex}
\titlespacing*{\subsection}{0pt}{0.6ex plus .2ex minus .2ex}{0.3ex plus .1ex}
\titlespacing*{\paragraph}{0pt}{0.4ex plus .1ex}{0.6em}
\title{Modal Logic Neural Networks}

\author{%
 \Name{Antonin Sulc} \Email{asulc@lbl.gov}\\
 \addr Lawrence Berkeley National Laboratory, Berkeley, CA, USA
 \AND
 \Name{Noor Naddour} \Email{n.naddour@uq.edu.au}\\
 \addr The University of Queensland, Brisbane, QLD, Australia
}

\begin{document}

\maketitle

\begin{abstract}
Neural Networks are indispensable to natural sciences and society. Their impact extends from applications in public health to workforce productivity. Here, we introduce Modal Logic Neural Networks (MLNNs) -- an end-to-end differentiable logical neural network realisation of modal logic which evaluates a learnable truth function across possible-world semantics. This neural architecture handles para-consistency and inconsistency via a learnable world accessibility relation and valuation function. Because the modality is fixed by which frame axioms the relation satisfies rather than by the operator, one differentiable engine covers the epistemic, doxastic, deontic and temporal readings, with applications from verification of reactive and distributed systems to legal discourse and microeconomic utility models. In this paper, we introduce a model of differentiable Kripke semantics, and establish their soundness, convergence, and structural guarantees. We show four applications, in which the learned relation reads as a trust matrix, an operating-regime embedding with safety bounds, a temporal precedence order, and a recovered constraint graph.

\end{abstract}


\section{Introduction}

\label{sec:intro}

While modern neural networks, especially large language models, primarily rely on statistical plausibility to generate outputs, Neurosymbolic AI combines connectionist learning and symbolic reasoning to achieve more human-like reasoning abilities. This addition of
differentiable logical constraints to neural architectures has been predominantly using classical logic (DeepProbLog~\cite{manhaeve2018deepproblog}, Semantic Loss~\cite{xu2018semantic},
Scallop~\cite{li2023scallop}). Modal logic, however, has long been used to reason about knowledge, belief, time and obligation, so a modal neural architecture would extend what such systems can express and generalise previous neurosymbolic architectures. Modal logic lets constraints carry modalities, for example a temporal safety rule ($\Box(\text{red}\to\neg\text{move})$), a deontic obligation ($\Box\phi_{\textsc{safe}}$), epistemic knowledge ($K_a\phi$) or a doxastic belief ($B_a\phi$). Truth values are propagated across a learnable Kripke structure, and inconsistencies are detected rather than hidden.

This paper introduces Modal Logic Neural Networks (MLNNs)
\footnote{\label{fn:code}Code and extended version available at \url{https://github.com/sulcantonin/torchmodal} (\texttt{pip install torchmodal}).}, a differentiable instantiation of Kripke semantics in which the worlds, the valuation, and the accessibility relation $A_\theta \in [0,1]^{|W| \times |W|}$ are all part of one forward pass, with the valuation, the relation, or both learnable, and truth carried as a bounded interval $[L,U]$ rather than a scalar (\L{}ukasiewicz operators relax $\min/\max$ to $\operatorname{smin}_\tau/\operatorname{smax}_\tau$; Figure~\ref{fig:architecture}, formalised in \S\ref{sec:method}).

\paragraph{Contributions.} \textbf{(1)}~A differentiable, end-to-end framework of a learnable world accessibility relation $A_\theta$ and valuation $V_\varphi$. The accessibility relation can be \textbf{T} reflexive , \textbf{4} transitive, \textbf{B} symmetric, \textbf{5} Euclidean, \textbf{D} serial (no dead ends). These properties are denoted as frame axioms, and are measurable after training so need not be imposed on the Kripke structure.
\textbf{(2)}~Interpretability built into inference, in two modes: deductive and inductive. We demonstrate both on four real-world examples whose outputs are domain-specific and human-interpretable. \textbf{(3)}~An open-source framework, called \textsf{torchmodal}, based on PyTorch\footref{fn:code}. \textsf{torchmodal} ships modal logic operators, accessibility relation parameterisations, and the frame-axiom audit, together with the four examples. \textbf{(4)}~A classical logic neural network as the zero-one special case of MLNNs, making MLNNs a generalisation encompassing classical LNNs.

\definecolor{slate900}{HTML}{0F172A}
\definecolor{slate700}{HTML}{334155}
\definecolor{slate400}{HTML}{94A3B8}
\definecolor{teal700}{HTML}{0F766E}
\definecolor{teal50}{HTML}{F0FDFA}
\definecolor{rose700}{HTML}{BE123C}
\definecolor{rose50}{HTML}{FFF1F2}
\definecolor{amber600}{HTML}{D97706}
\definecolor{amber50}{HTML}{FFFBEB}
\definecolor{slate50}{HTML}{F8FAFC}
%, unified figure palette (all figures use these; Seaborn-deep base),
\definecolor{mlblue}{HTML}{4C72B0}    % primary / heat-map ramp base
\definecolor{mlamber}{HTML}{DD8452}   % secondary
\definecolor{mlgreen}{HTML}{55A868}   % tertiary / correct / true edge
\definecolor{mlred}{HTML}{C44E52}     % negative / wrong / violation
\definecolor{mlpurple}{HTML}{8172B3}  % quaternary
\definecolor{mlgray}{HTML}{8C8C8C}    % neutral / satisfied edge

\begin{figure}[t]
\centering
\resizebox{\textwidth}{!}{%
\begin{tikzpicture}[
    font=\sffamily,
    >=Stealth,
    world/.style={circle, draw=teal700, line width=0.9pt, fill=teal50,
        minimum size=1.55cm, text=teal700, font=\LARGE},
    embed/.style={circle, draw=teal700, line width=0.7pt, fill=white,
        minimum size=0.85cm, text=teal700, font=\normalsize},
    val_box/.style={rectangle, draw=teal700, line width=0.6pt, fill=teal50!50,
        rounded corners=1.0mm, minimum width=2.4cm, minimum height=0.78cm,
        font=\large, align=center},
    learn_strong/.style={->, line width=1.7pt, draw=rose700,
        shorten >=2pt, shorten <=2pt},
    learn_mid/.style={->, line width=1.05pt, draw=rose700!75,
        shorten >=2pt, shorten <=2pt},
    learn_weak/.style={->, dashed, semithick, draw=rose700!45,
        shorten >=2pt, shorten <=2pt},
    panel_label/.style={font=\Large\bfseries, text=slate700},
    encode/.style={->, line width=1.1pt, dashed, draw=slate700!60},
]

% ---------- (a) LEFT: reasoning over worlds ----------
\node[panel_label] at (3.0, 4.7) {Reasoning over worlds};
\node[world] (w1) at (1.4, 2.7) {$w_1$};
\node[world] (w2) at (4.6, 2.7) {$w_2$};
\node[world] (w3) at (3.0, 0.9) {$w_3$};
\node[val_box, anchor=south, yshift=2mm] (vw1) at (w1.north) {$V_{w_1}(\phi){=}0.9$};
\node[val_box, anchor=south, yshift=2mm] (vw2) at (w2.north) {$V_{w_2}(\phi){=}0.4$};
\node[val_box, anchor=north, yshift=-2mm] (vw3) at (w3.south) {$V_{w_3}(\phi){=}0.7$};
\draw[learn_strong] (w1) -- node[above=0.5mm, font=\large, text=rose700] {$A_\theta{=}0.95$} (w2);
\draw[learn_mid]    (w1) -- node[left=1mm, font=\large, text=rose700!85] {$A_\theta{=}0.62$} (w3);
\draw[learn_weak]   (w2) -- node[right=1mm, font=\large, text=rose700!55] {$A_\theta{=}0.05$} (w3);
\draw[learn_strong] (w1) edge[in=200, out=160, looseness=14]
    node[left=0.5mm, font=\large, text=rose700] {$A_\theta{=}1.00$} (w1);
\node[font=\normalsize\itshape, text=slate700] at (3.0, -1.45)
    {aggregating at $w_1$ brackets $V_{w_1}$: $[\Box\phi,\Diamond\phi]{=}[0.45,\,0.90]$};

% ---------- encoder bridge ----------
\draw[encode] (6.10, 2.05) -- node[above=1pt, font=\Large\itshape, text=slate700] {$f_\theta$} (7.40, 2.05);

% ---------- (b) RIGHT: worlds in the manifold ----------
\begin{scope}[xshift=7.7cm]   % <<< panel gap: adjust this one number
\node[panel_label] at (3.0, 4.7) {Worlds in the manifold};
\fill[slate50]
    plot[smooth cycle, tension=0.7] coordinates
    {(-0.05,0.4) (1.4,-0.3) (3.5,-0.35) (5.4,0.25) (6.3,1.3)
     (6.2,2.7) (5.0,3.5) (3.0,3.7) (1.0,3.4) (0.0,2.2) (-0.25,1.0)};
\fill[rose700!11]
    plot[smooth cycle, tension=0.65] coordinates
    {(0.35,0.55) (1.6,0.0) (3.5,-0.05) (5.2,0.5) (5.95,1.5)
     (5.75,2.85) (4.5,3.35) (2.6,3.45) (0.85,3.1) (0.15,2.0) (-0.05,1.1)};
\fill[teal700!18]
    plot[smooth cycle, tension=0.7] coordinates
    {(0.85,1.0) (1.85,0.5) (3.5,0.55) (4.8,1.0) (5.2,1.85)
     (4.95,2.85) (3.65,3.15) (2.0,3.05) (0.95,2.55) (0.55,1.65)};
\fill[teal700!32]
    plot[smooth cycle, tension=0.7] coordinates
    {(1.4,1.4) (2.3,1.0) (3.6,1.05) (4.35,1.5) (4.55,2.45)
     (3.5,2.9) (2.0,2.85) (1.3,2.2)};
\fill[teal700!55]
    plot[smooth cycle, tension=0.7] coordinates
    {(1.95,1.85) (2.65,1.65) (3.4,1.85) (3.4,2.4) (2.65,2.65) (1.9,2.45)};
\node[font=\large\bfseries, text=white] at (3.05, 2.55) {$\Box\phi$};
\node[font=\normalsize\itshape, text=teal700!90] at (4.55, 1.55) {$\phi$ holds};
\node[font=\normalsize\itshape, text=rose700!85] at (5.30, 2.85) {$\Diamond\phi$ reachable};
\node[embed] (h1) at (2.45, 2.05) {$h_{w_1}$};
\node[embed] (h2) at (4.05, 2.20) {$h_{w_2}$};
\node[embed] (h3) at (1.05, 1.20) {$h_{w_3}$};
% \node[font=\normalsize\itshape, text=slate700] at (3.0, -1.45)
%    {$\Box$ shrinks a region, $\Diamond$ expands it};
\end{scope}
\end{tikzpicture}}
\caption{\textbf{Modal Logical Neural Networks.} \textbf{Left:} three worlds, each with its own truth value $V_w(\phi)$, joined by learned accessibility weights $A_\theta\in[0,1]$ saying how much one world's verdict bears on another's. The weights are the parameters. One modal neuron firing at $w_1$ (Eq.~\ref{eq:modal_operators}) gives necessity $\Box\phi(w_1){=}\operatorname{smin}_\tau[(1{-}A_\theta){+}V]{\approx}0.45$ (``how true in \emph{every} world $w_1$ sees'') and possibility $\Diamond\phi(w_1){=}\operatorname{smax}_\tau[A_\theta{+}V{-}1]{\approx}0.90$ (``in \emph{some}''). They bracket $V_{w_1}{=}0.9$, so the bound is consistent and $L_{\text{contra}}{=}0$; had they crossed, that gap is the training signal. \textbf{Right:} the encoder $f_\theta$ turns worlds into points $h_w$.}
\label{fig:architecture}
\end{figure}
\section{Related Work}

\paragraph{Neurosymbolic AI.}
DeepProbLog~\cite{manhaeve2018deepproblog} combines neural networks with
probabilistic logic programming; the Semantic Loss~\cite{xu2018semantic}
enforces propositional constraints on neural outputs, functionally close to our
$L_{\text{contra}}$ but single-world; Scallop~\cite{li2023scallop} and
NeurASP~\cite{yang2020neurasp} give differentiable probabilistic Datalog and
ASP; Logic Tensor Networks~\cite{badreddine2022ltn} ground first-order logic
in real-valued semantics.

\paragraph{Modal Logic in AI.}
Modal logic has a long history in AI for reasoning about knowledge, belief, time, and obligation~\cite{fagin1995reasoning,hintikka1962knowledge,liau2003belief}, and supplies the logical foundations of multiagent systems~\cite{shoham2009multiagent}. Symbolic tools such as mlsolver~\cite{rohkohl2024mlsolver} build Kripke structures and check axiom sets for contradictions or tautologies before deployment; MLNN is complementary, fitting the accessibility relation to data and exposing it as the inspectable forward-pass artefact rather than requiring it to be specified in advance.

\paragraph{Fuzzy modal logic and smooth min/max.} Many-valued and {\L}ukasiewicz modal logics with fuzzy accessibility are established~\cite{fitting1991mvml,bou2011mvmodal,hajek2001metamathematics,caicedo2010standard} and are essentially the semantics our forward pass computes; what is new is that the frame is a \emph{learned parameter} rather than a fixed structure. 

\paragraph{Logical and Modal Neural Networks.}
LNNs~\cite{riegel2020logical} introduced weighted real-valued logic with provable bounds for the propositional case; we extend the re-grounding to modal logic by making the Kripke triple $(W, R, V)$ an explicit, either-fixed-or-learnable component, with $\Box / \Diamond$ realised as bounded sound aggregators rather than recurrent transition operators. Connectionist Modal Logic~\cite{Garcez2007CML} embeds modal operators in recurrent hidden states and, like MLNN, is trained end-to-end, but its accessibility relation is entangled in those hidden states and never materialised; our $A_\theta$ is an explicit, inspectable first-class object. STLCG~\cite{leung2023stlcg} addresses Signal Temporal Logic via robustness values; our $[L,U]$ formulation separates epistemic uncertainty from degree of satisfaction and generalises beyond fixed temporal ordering. Recent work on temporal-logic satisfiability~\cite{Luo2022LTLfNet} and epistemic logic in RL~\cite{Engesser2025EpistemicRL} addresses specific modal fragments; in contrast, MLNN provides a single, end-to-end-differentiable framework in which the accessibility relation itself is learned by gradient descent. The \L{}ukasiewicz choice for Boolean connectives follows~\cite{vankrieken2022fuzzy}; MLNs~\cite{richardson2006markov} and ProbLog~\cite{deraedt2007problog} combine first-order logic with probability but do not natively support modal operators. Differentiable constraint solvers such as SATNet~\cite{wang2019satnet} (a learnable MaxSAT layer) and recurrent relational networks~\cite{palm2018rrn} solve combinatorial problems by learning the constraint rules from a corpus of solved instances; MLNN instead solves each instance directly from explicit modal axioms and additionally learns the accessibility relation itself, a complementarity we make precise on graph colouring (\S\ref{sec:exp:coloring}). In several instantiations (C-MAPSS, ROCStories) this relation is metric-learned from an embedding $f_\theta$, an additional interpretability layer atop the provable frame axioms; both the valuation $V$ and the accessibility $A_\theta$ can be learned, differentiable, and inspected.

\section{Method: Modal Logic Neural Networks}
\label{sec:method}

\subsection{MLNN Object Language}

\label{sec:method:syntax}

The MLNN supports the following grammar of logical formulae:
\begin{equation}
\phi ::= p \mid \neg \phi \mid \phi \land \psi \mid \phi \lor \psi \mid \phi \to \psi \mid \Box \phi \mid \Diamond \phi \mid \phi \mathbin{\mathcal{U}} \psi
\label{eq:grammar}
\end{equation}
where $p$ ranges over atomic propositions. We define the set of well-formed formulas inductively over the formulas in \ref{eq:grammar} above.
Boolean connectives use {\L}ukasiewicz fuzzy logic~\cite{vankrieken2022fuzzy}: $\neg \phi = 1 - \phi$; $\phi \land \psi = \max(0, \phi + \psi - 1)$; $\phi \lor \psi = \min(1, \phi + \psi)$; $\phi \to \psi = \min(1, 1 - \phi + \psi)$.
Each formula $\phi$ is represented by truth bounds $[L_\phi, U_\phi] \subseteq [0,1]$ per world, forming a DAG-structured formula graph where nodes are subformulae and edges are dependencies. Inference propagates bounds through this graph using the upward-downward bound-propagation algorithm of LNNs~\cite{riegel2020logical}, detailed in the extended version of the paper~\footref{fn:code}.
% detailed in the supplementary material (\emph{Complete Theoretical Analysis}).

\subsection{Kripke Semantics}
\label{sec:method:prelimintro}

A modal Kripke frame $F=(W,R)$ consists of a non-empty set (of worlds) $W$, and a binary world accessibility relation $R \subseteq W \times W$. A valuation $V$ of the object language on a frame $F=(W,R)$ is a map associating with each proposition $p$ a set of worlds $V(p)$. This is understood as the set of worlds at which $p$ is true. A Kripke model of is a pair $M=(F,V)$ where $F$ is a frame and $V$ is a valuation on $F$. We define a truth-relation $(M,x)\vDash \phi$ where $\phi$ is a well-formed formula of the object language to mean formula $\phi$ is true at world $x$ in model $M$. A formula is true in a model $M=(F,V)$ if for every $w \in W$, $w \vDash \phi$. A formula $\phi$ is valid in a frame $F$ if $\phi$ is true in all models based on $F$, in symbols $F \vDash \phi$. We define modal logic $K$ in the object language as the set of all formulas that are valid in all modal Kripke frames.

% A Kripke model is a tuple $M=\langle W,R,V\rangle$: $W$ \nn{is} a finite set of possible
% worlds, $R\subseteq W\times W$ \nn{is a binary relation, called accessibility relation} a binary accessibility relation, and $V$ \nn{is} a
% valuation \nn{a function from the atoms to a subset of $W$} giving each atomic $p$ a truth value in each world.
MLNN is a differentiable, learnable instance of this, where a ``world'' may be an agent's
belief state, a moment in time, or any context. Truth values become bounds
$[L,U]\subseteq[0,1]$, and the inter-world structure is either a fixed crisp $R$ or a learnable $A_\theta\in[0,1]^{|W|\times|W|}$, optionally sparsified by
top-$k$ masking to $\tilde{A}$ (\emph{Masking and Sparsity} in the extended version\footref{fn:code}). The
operators $\Box$ (necessity) and $\Diamond$ (possibility) can have any modality interpretation; epistemic
($K_a$), doxastic ($B_a$), temporal ($G/F$) or deontic ($O/P$). The interpretation is fixed by the user. With \textsf{torchmodal} only one computational engine is needed. The user specifies the frame axioms $A_\theta$ as needed. For an example, one can specify a reflexive and transitive frame and obtain $S4$ modal Kripke model.
% Only knowledge assumes reflexivity (axiom T), and that is what makes
% it factive: what is known is true. Belief and obligation do not, so what is
% believed or obligatory need not hold: giving S5, KD45 and KD.
Temperature $\tau$ controls the soft aggregations; training minimises $L_{\text{task}}$
and the contradiction loss $L_{\text{contra}}$, balanced by $\beta$.

\subsection{Differentiable Kripke Semantics}

Each Kripke component becomes a differentiable tensor. For proposition $p$ the
MLNN stores truth bounds of shape $(|W|,2)$, each row $[L_{p,w},U_{p,w}]$ giving
the bounds of $p$ in world $w$. A scalar valuation conflates two distinct
uncertainties, how strongly evidence supports $p$ and how strongly it rules $p$
out; the pair separates them ($L$ guarantees how true $p$ is, $U$ how
not-false), with scalars the degenerate case $L{=}U$. Two reasons to carry both.
\emph{(i)} Many evidence sources emit two numbers: an NLI head returns separate
entailment and contradiction probabilities, and setting $L{=}\text{entail}$,
$U{=}1-\text{contradict}$ lets an internally inconsistent source surface
directly as $L>U$. \emph{(ii)} The same shape expresses the bound contradiction
$\max(0,L_\phi-U_\phi)$, an inconsistency no classical assignment satisfies.

\subsubsection{Modal Operators: The Necessity and Possibility Neurons}

The modal operators are neurons that aggregate across worlds. Classical logic
takes hard $\min$/$\max$ over a fixed neighbourhood; we use differentiable
relaxations over the weighted $\tilde{A}$ so gradients pass through the
structural decision. For truth values $x=\{x_i\}$,
$\operatorname{smin}_\tau(x)=-\tau\log\sum_i\exp(-x_i/\tau)$ is a sound lower
bound on $\min(x)$ and $\operatorname{smax}_\tau(x)=\tau\log\sum_i\exp(x_i/\tau)$
a sound upper bound on $\max(x)$;\footnote{$\operatorname{smin}$/$\operatorname{smax}$
distinguish these log-sum-exp aggregators from the probability-normalising
$\mathrm{softmax}$.} and $\operatorname{conv\text{-}pool}_\tau(x,z)=\sum_i w_ix_i$
with $w_i=\mathrm{softmax}(z_i/\tau)$ is a convex pool, lower-bounding $\max(x)$
at $z{=}x$ and upper-bounding $\min(x)$ at $z{=}-x$. Default $\tau=0.1$.

\paragraph{The $\Box$ (Necessity) and $\Diamond$ (Possibility) Neurons}
%$\Box\phi$ are a weighted universal quantification (``weakest link detector'') and $\Diamond\phi$ a weighted existential (``evidence scout''). 
$\Box\phi$ is a weighted universal quantifier (``weakest link detector'') and $\Diamond\phi$ a weighted existential one (``evidence scout'').
With $\bar{A}_{w,w'}{:=}1-\tilde{A}_{w,w'}$,
{\small
\begin{equation}
\begin{aligned}
L_{\Box\phi,w} &= \underset{w'}{\operatorname{smin}_\tau}\!\bigl(\bar{A}_{w,w'}{+}L_{\phi,w'}\bigr), &
U_{\Box\phi,w} &= \underset{w'}{\operatorname{conv\text{-}pool}_\tau}\!\bigl(\bar{A}_{w,w'}{+}U_{\phi,w'}\bigr),\\[2pt]
L_{\Diamond\phi,w} &= \underset{w'}{\operatorname{conv\text{-}pool}_\tau}\!\bigl(\tilde{A}_{w,w'}{+}L_{\phi,w'}{-}1\bigr), &
U_{\Diamond\phi,w} &= \underset{w'}{\operatorname{smax}_\tau}\!\bigl(\tilde{A}_{w,w'}{+}U_{\phi,w'}{-}1\bigr).
\end{aligned}
\label{eq:modal_operators}
\end{equation}
}%
Each $\operatorname{conv\text{-}pool}_\tau$ above takes the bounding direction of its definition: $z{=}{-}x$ for $U_{\Box\phi}$ (an upper bound on $\min$) and $z{=}x$ for $L_{\Diamond\phi}$ (a lower bound on $\max$), with $x$ the bracketed argument shown. The duality $\Diamond\phi \equiv \neg\Box\neg\phi$ holds via $\operatorname{smax}(x) = 1 - \operatorname{smin}(1-x)$ (derived in \emph{De Morgan Duality of the Modal Operators} in the extended version\footref{fn:code}). All four bounds in Equation~\ref{eq:modal_operators} are clamped to $[0,1]$ before use: the explicit clamp matters because $\operatorname{smin}_\tau$ can exceed $1$ on a single-term bracket, which would break the $[0,1]$ truth-bound invariant. 

\paragraph{The Until ($\mathcal{U}$) Neuron.}
When $A_\theta$ encodes a temporal order, the until $\phi \mathbin{\mathcal{U}} \psi$ (``$\phi$ holds until $\psi$ becomes true'') is a third modal aggregation. Over worlds in time order $1,\dots,T$ its bounds follow the backward {\L}ukasiewicz recurrence
$(\phi \mathbin{\mathcal{U}} \psi)_t = \psi_t \lor \bigl(\phi_t \land (\phi \mathbin{\mathcal{U}} \psi)_{t+1}\bigr)$, terminating at $(\phi \mathbin{\mathcal{U}} \psi)_T = \psi_T$, applied to the $L$- and $U$-bounds in turn (equivalently $\bigvee_{t' \ge t} \psi_{t'} \land \bigwedge_{t \le s < t'} \phi_s$); the sentence-ordering experiment (\S\ref{sec:exp:rocstories}) scores a single such until at the first world.

\paragraph{Handling Contradictions.}
The $[L, U]$ bound representation detects and localises contradictions in user-provided axioms rather than silently producing inconsistent outputs. Atomic propositions and non-modal compound nodes are constructed so that $L_{\phi,w} \leq U_{\phi,w}$ holds by construction. Modal compound nodes, however, can produce $L_{\phi,w} > U_{\phi,w}$ during the upward pass when the current accessibility relation $A_\theta$ is incompatible with the user-supplied axioms; this inversion is not projected back to a valid interval but surfaced as the per-formula, per-world residual $\max(0, L_{\phi,w} - U_{\phi,w})$ that contributes to $L_{\text{contra}}$. Gradient descent then resolves the inversion by adjusting the propositional content (deductive mode), the accessibility structure (inductive mode), or both. This contrasts with single-valued robustness approaches (e.g., STLCG), where conflicting constraints can cancel silently.

\subsection{Flexible and Learnable Accessibility Relations}
\label{sec:method:A}

Unlike classical modal logic, where $R$ is fixed, MLNN can treat the accessibility relation as a learnable parameter $A_\theta$. Both
modes are exercised: colouring uses a fixed adjacency when solving a given
instance, while C-MAPSS learns $A_\theta$ as a metric kernel and the inductive
colouring task recovers it from data alone (\S\ref{sec:experiments}). We
parameterise $A_\theta:W\times W\to[0,1]$ either as a matrix of learnable logits
through a sigmoid (small domains) or, for scale, by metric learning: an encoder
maps each world to $\mathbf{h}_w\in\mathbb{R}^d$ and
$A(w_i,w_j)=\sigma(\mathbf{h}_{w_i}^\top\mathbf{h}_{w_j})$ makes logical access
geometric proximity, reducing parameters from quadratic to linear in $|W|$. The
weights enter the modal neurons through the implication
$1-(\tilde A)_{ij}+L_{\phi,w_j}$ inside the $\operatorname{smin}$ of
Equation~\ref{eq:modal_operators}: at $(\tilde A)_{ij}\approx1$ the truth value
fully participates in the minimum, at $\approx0$ it is removed. Gradient descent
on the contradiction loss updates $\theta$, which is the inductive capability
that lets the model discover the structure best resolving the data's
contradictions.

\subsection{Loss Function and Training}

Training minimises $L_{\text{total}}$ end to end, where $L_{\text{task}}$ is any supervised loss and $L_{\text{contra}}$ penalises the crossed bounds that the modal neurons (Eq.~\ref{eq:modal_operators}) can produce:
{\small
\begin{equation}
L_{\text{total}} = L_{\text{task}} + \beta L_{\text{contra}},
\qquad
L_{\text{contra}} = \sum_{w \in W}\sum_{\phi} \max\!\left(0,\, L_{\phi,w} - U_{\phi,w}\right).
\label{eq:ltotal}
\end{equation}
}%
\noindent\phantomsection\label{eq:contra}%
\noindent where $\beta$ trades task accuracy against logical coherence. For pure
constraint satisfaction we set $L_{\text{task}}{=}0$ and learn entirely from
$L_{\text{contra}}$ (\emph{pure satisfiability mode}), making the MLNN a
differentiable energy-based model whose free parameters are the proposal logits.

\section{Theoretical Properties}
\label{sec:theory}

We state the properties below in brief; full proofs are in \emph{Complete Theoretical Analysis} in the extended version\footref{fn:code}.
% The properties below are shown in full in the supplementary material under \emph{Complete Theoretical Analysis}:

\noindent \emph{Soundness.} An aggregator is \emph{sound} when it never claims more than
the operator it relaxes, so its output brackets that $\min/\max$ and a satisfied
axiom under the soft reading never gives false reassurance about the crisp one:
$\operatorname{smin}_\tau(x)\leq\min(x)$ and
$\operatorname{conv\text{-}pool}_\tau(x,-x)\geq\min(x)$ for $x_i\in[0,1]$, with
equality as $\tau\to0$. This bounds the operators, not the graph: a modal node
may still emit $L_\phi>U_\phi$, the contradiction signal described above.

\noindent \emph{Convergence.} For acyclic graphs over the non-modal fragment, with
polarity-tracked monotone operators and fixed parameters, each sweep is exact at
$O(|V|+|E|)$ operator evaluations; \emph{acyclicity} makes the count finite, the
monotone-bounded argument alone giving only asymptotic convergence. The
\emph{joint} fixed point needs more than one round whenever axioms share a
subformula, a downward update stales a sibling already consumed upward, so
inference iterates to a tolerance. Cyclic graphs get no finite bound.

\noindent \emph{Structural recoverability.} For each of T, 4, B, 5, D a differentiable
regulariser $R_{\mathcal{P}}\ge0$ vanishes exactly on relations with that
property, and adding it preserves soundness. Four qualifications.
\emph{(i)} For four of the five this is well-posedness, not a training-dynamics
guarantee: $R_{\mathcal{P}}$ is the property's residual, so its zero set is the
property by construction; only seriality has content beyond the definition.
\emph{(ii)} {No experiment here uses them} and all four run in the
{audit} regime, axioms measured post-hoc, so reported values are emergent
and can fail silently, which is why we measure them. \emph{(iii)} Exact
satisfaction is unattainable under $A_\theta=\sigma(\widetilde{A}_\theta)$, whose
entries lie in the \emph{open} $(0,1)$: an exact $1.00$ signals saturation or a
kernel symmetry, not a crisp relation. \emph{(iv)} The regularisers use the
{product} t-norm and the audit {\L}ukasiewicz; the former is strictly
stronger, so $R_{\mathcal{P}}{=}0$ implies the audit passes but not conversely.

\noindent \emph{Complexity.} A modal neuron costs $O(|W|)$ under a fixed sparse relation and $O(|W|^2)$ dense; the metric parameterisation (\S\ref{sec:method:A})
defines accessibility over $d$-dimensional embeddings ($d\ll|W|$), cutting
parameters to $O(d\cdot|W|)$ and, with top-$k$ masking, compute to
$O(k\cdot|W|)$ per neuron. Forming that mask exactly still scores every pair, so end-to-end cost stays $O(|W|^2)$, as measured in \emph{Scalability Analysis} in the extended version\footref{fn:code}: the parameter
saving removes the dense cliff, not the exponent.

\section{Experiments}\label{sec:experiments}

We evaluate MLNN on four examples spanning the modal range and both directions
of use. 
Our claim is that the relation the model uses to make predictions is the same relation an analyst can inspect, and that this comes at accuracy competitive with strong task-specific baselines, which we report alongside. The contribution is this inspectable modal structure, which the baselines do not provide.
One inference engine serves all four, every one of them an instance of Equation~\ref{eq:ltotal} under contradiction loss alone (C-MAPSS, colouring, ROCStories, doxastic). Each answers one question: whether a learned trust relation exposes an incoherent \emph{stance}, and whether the modal layer adds anything its scorer does not (\S\ref{sec:exp:doxastic}); whether one relation trained on a deontic contradiction supports several independent analyses at once (\S\ref{sec:exp:cmapss}); whether a modal operator matters or a propositional scorer on the same embeddings would do as well (\S\ref{sec:exp:rocstories}); and, where ground truth for the relation exists,
whether the learned relation is the correct one, i.e.\ whether the artefact is
faithful rather than merely plausible (\S\ref{sec:exp:coloring}).

\vspace{-0.35em}\subsection{Doxastic Deception: a trust matrix that surfaces a saboteur the LLM-judge misclassifies}\label{sec:exp:doxastic}

The \emph{doxastic} reading $B_a\phi$ (``agent $a$ believes $\phi$'') audits not
what is true but whether an agent's stance coheres: a multi-agent LLM system can
be derailed by a member holding an incoherent belief without ever making a
checkably false claim. Six agents, each the same locally-served LLM (via
\textsf{ollama}), negotiate a \$$50$K budget over twelve rounds; agent~$2$ is a
saboteur. A frozen NLI head scores every utterance against ``the budget is
fair'' into a per-round (entail, contradict), summarised per agent as a bound
$[L_j,U_j]$; a persistent flip-flopper drives $L_j>U_j$, a crossed bound no
truth value holds. A multi-agent Kripke model trains the $6{\times}6$ trust
matrix by maximising $\Box_i\mathrm{coherent}$ over the trust relation (the
doxastic axiom $A_\theta[i,j]\!\to\!\mathrm{consistent}(j)$,
$\mathrm{consistent}(j){=}1-\mathrm{relu}(L_j-U_j)$), so a
self-contradictory agent collapses its column toward zero and the predicted
saboteur is $\arg\min_j\overline{A_\theta[\cdot,j]}$. The baseline re-prompts
the same LLM with the raw transcript to name the saboteur.

\vspace{-0.35em}\paragraph{Result and interpretation.} We run five backbones $\times$ three
seeds per scenario, so every cell below is a detection count out of $15$, and we
report a Wilson $95\%$ interval with it. Because MLNN and the judge read the
\emph{same} transcript, the two are paired and we test them with an exact
McNemar test rather than a two-proportion test. Three regimes appear. When the
saboteur flip-flops blatantly both tie ($15/15$, CI $[0.80,1.00]$): the
transcript is obvious. The separation is the subtle case, where the saboteur
drifts politely while a stylistically loud but content-honest agent looks
suspicious. The LLM-judge, reading surface form, identifies the loud-honest decoy and scores $3/15$ ($[0.07,0.45]$); MLNN, whose NLI front-end scores entailment
rather than sentiment, discounts the tone and reaches $12/15$ ($[0.55,0.93]$;
McNemar $p=0.004$, $9$ discordant pairs, all favouring MLNN). Under prompt
\emph{injection}, where the saboteur embeds a clause framing an honest agent,
MLNN detects the saboteur in $15/15$ runs versus $5/15$ ($p=0.002$): the injection text lives in the saboteur's own transcript and cannot move the framed agent's column, an immunity the modal pipeline has by construction rather than by training.

\begin{figure}[t]
\centering
\begin{minipage}[t]{0.32\linewidth}\centering
  \includegraphics[width=\linewidth]{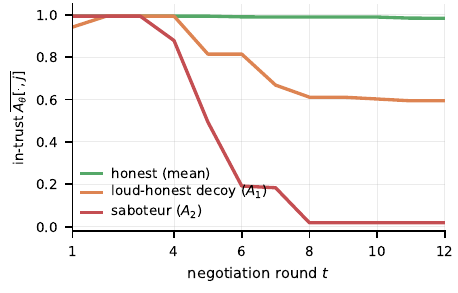}\\[-2pt]
  {\footnotesize \textbf{(a)} trust per round}
\end{minipage}\hfill
\begin{minipage}[t]{0.32\linewidth}\centering
  \includegraphics[width=\linewidth]{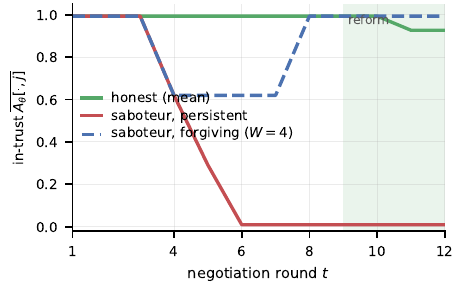}\\[-2pt]
  {\footnotesize \textbf{(b)} a reforming saboteur}
\end{minipage}\hfill
\begin{minipage}[t]{0.32\linewidth}\centering
  \includegraphics[width=\linewidth]{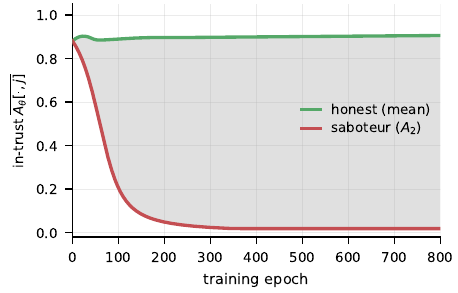}\\[-2pt]
  {\footnotesize \textbf{(c)} training convergence}
\end{minipage}
\caption{\textbf{The object the modal layer produces} (\textsf{llama3}): the
in-trust $\overline{A_\theta[\cdot,j]}$ others accord each agent.
\textbf{(a)} Subtle, per round --- the saboteur drains to $0.018$ from round~$4$,
honest agents hold near $0.98$, the decoy, penalised only for tone, settles at
$0.59$; detection evaluates the top-$3$ mean, not this curve.
\textbf{(b)} a saboteur backing the budget from round~$9$ (shaded): the default
\emph{persistent} summary stays collapsed at $0.009$, a \emph{forgiving} $W{=}4$
window lets trust return: one knob, same pipeline.
\textbf{(c)} training: the gap reaches $90\%$ of its final $0.89$ by epoch~$142$.}
\label{fig:doxastic}
\end{figure}

\vspace{-0.35em}\paragraph{Role of modal layer.} The pipeline stacks three components and the gap above could belong to any of them, so we peel them apart on identical transcripts (\emph{Doxastic Deception} in the extended version\footref{fn:code}). The
frozen NLI head \emph{alone}, with no bound, no trust matrix and no training,
already reaches $15/13/15$ against the full stack's $15/12/15$, a difference
significant in no scenario (McNemar $p=1.0$). The detection gap over the LLM-judge is therefore due to the NLI front-end, not the modal layer, and a judge given the same front-end would close it. What
the modal layer contributes is the object behind the verdict, a per-cell trust
matrix with the utterances that drained each column
(Figure~\ref{fig:doxastic}), and the structural injection-immunity above, which
follows from per-agent column isolation and which no transcript-level scorer
has. That claim is narrower than a detection win, and all three scenarios are
author-designed, so these numbers measure detection of a constructed failure
mode rather than its prevalence in the wild.

\vspace{-0.35em}\subsection{Turbofan Wear (C-MAPSS): one objective, two views and a control}\label{sec:exp:cmapss}

C-MAPSS~\cite{cmapss2008} is NASA's standard run-to-failure benchmark for
turbofan engines; FD002 runs each engine from healthy to failure under six
unlabelled operating regimes, recording $21$ sensors, $3$ operating settings and
a ground-truth Remaining Useful Life (RUL) per cycle. Each (engine, cycle) pair
is a Kripke world, and both Kripke components are learned from the
$24$-dimensional sensor state alone: a metric head gives the within-engine
accessibility $A_\theta(w_i,w_j)=\sigma(s\,f_\theta(s_i)^\top f_\theta(s_j)+b)$
with $f_\theta:\mathbb{R}^{24}\!\to\!\mathbb{R}^{16}$ ($2642$ parameters), and a
second head the valuation $V_\phi(s)=\sigma(\mathrm{MLP}(s))$ ($1665$
parameters), $V_{\textsc{safe}}=1-V_\phi$. With deontic predicates
$\textsc{safe}=(\mathrm{RUL}{>}100)$ and
$\textsc{fail-soon}=(\mathrm{RUL}{<}30)$, the constraint determines the per-cycle
bounds $L_{\Box(\textsc{safe})}$ and $U_{\Diamond(\textsc{fail\text{-}soon})}$.
The heads carry disjoint gradients: $\mathcal{L}_V=\mathrm{BCE}(V_\phi(s),
\mathbb{1}[\mathrm{RUL}{<}30])$ trains the valuation, while $A_\theta$ is
trained by the deontic contradiction alone, with $V$ detached,
\begin{equation}
\mathcal{L}_{\mathrm{deon}} =
\underbrace{\big(1-U_{\Diamond(\textsc{fail-soon})}\big)^2}_{\mathrm{RUL}<30}
\;+\;
\underbrace{U_{\Diamond(\textsc{fail-soon})}^2+\big(1-L_{\Box(\textsc{safe})}\big)^2}_{\mathrm{RUL}>100},
\label{eq:deontic-contra}
\end{equation}
the at-risk middle carrying no term and being shaped only through $A_\theta$. No
locality kernel or other constructed target is fitted. RUL supervises $V_\phi$
during training but is never read at inference: the safety envelope is therefore
not an unsupervised read, whereas the regime and degradation probes are, since
no loss references either.

\begin{figure}[t]
\centering
\includegraphics[width=0.92\linewidth]{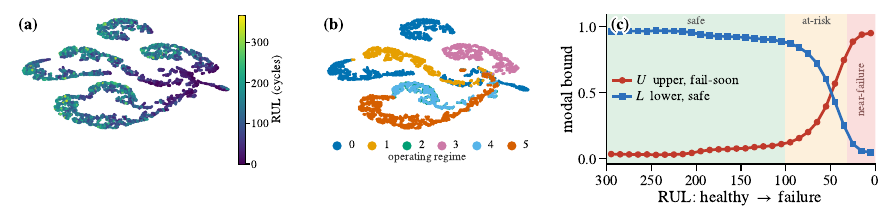}
\caption{\textbf{Turbofan Wear: three analyses of one relation trained only on the deontic contradiction.} \textbf{(a)} the embedding $f_\theta$ (t-SNE) coloured by
RUL: worlds organise along a wear manifold. \textbf{(b)} the same embedding, same
axes, coloured by operating regime. \textbf{(c)} the deontic bounds against RUL.
Both $V_\phi$ and $A_\theta$ read sensors only; no RUL is read at inference.
Panels are seed $45$, quoted statistics over $3$ seeds.}
\label{fig:results_combined}
\end{figure}

\vspace{-0.35em}\paragraph{Result and interpretation.} 
This is the deontic case, and the point is applicability, not accuracy. From one rule where safe must hold, fail-soon may, never both and the model learns a per-cycle safety margin that tracks wear, extends to unlabelled operating states, and matches a standard classifier while staying inspectable. On the whole FD002 fleet ($260$ train /$259$ test engines, $48{,}299$ worlds; $3$ seeds, mean${\pm}$std), one contradiction objective supports two analyses and a control (Figure~\ref{fig:results_combined}).
\emph{(1) Regimes.} $f_\theta$ retains the six operating regimes although no
loss named them: linearly decodable at $0.911{\pm}0.008$, and canonically
correlated with the raw settings at $\rho\!\approx\!0.88$. 
This is preservation rather than discovery: the regime label is $k$-means on three columns of the encoder's own input, and a linear probe on that input already reaches 0.929 (majority class 0.301). The result therefore shows that the deontic objective leaves the regime partition almost intact under a $24\to16$ bottleneck.
\emph{(2) Degradation.} The same embedding carries the wear axis although only
the $\mathrm{RUL}{<}30$ indicator was supervised and a ridge probe to continuous RUL reaches $\rho=0.710{\pm}0.007$ where the bounds follow it,
$U_{\Diamond(\textsc{fail\text{-}soon})}$ rising $0.07\!\to\!0.34\!\to\!0.93$ by
state as $L_{\Box(\textsc{safe})}$ erodes $0.93\!\to\!0.66\!\to\!0.07$.
\emph{(3) Frame structure.} The audit scores the relation as reflexive, symmetric and transitive (T$=0.896\pm0.007$, B$=1.000$, 4$=0.998\pm0.001$). These scores come from the parameterisation, not from learning: an inner-product kernel is symmetric by design and gives every world high access to itself, and the relation is so sparse that the transitivity test almost never applies. Compared with randomly shuffled relations of the same shape (\emph{Frame-axiom audit} in the extended version\footref{fn:code}), no axiom scores meaningfully higher.
% The relation audits as reflexive and symmetric, and nominally transitive ($\mathrm{T}{=}0.896{\pm}0.007$, $B{=}1.000$, $4{=}0.998{\pm}0.001$), but this is the parameterisation, not a learned frame: the accessibility is an inner-product kernel, so it is symmetric and strongly self-accessing by construction, and the implication axioms are near-vacuous. Under the diagonal-preserving null of the supplement, no axiom shows a meaningful gap.

Held-out detection is reported for completeness, not as a competitive claim: the upper bound scores AUROC $0.980{\pm}0.001$ against $0.990$ for its own valuation head and $0.988$ for logistic regression on the same sensors. The modal aggregation does not improve on the classifier it smooths; what it adds are the structural insights that (1) and (2) show. %Per-state numbers and the five-axiom audit are in the supplement (\emph{Turbofan Wear}).

\vspace{-0.35em}\subsection{Sentence Ordering (ROCStories): a learned temporal precedence relation}\label{sec:exp:rocstories}

The temporal modality exercises the \emph{Until} operator $\mathcal{U}$, the one
modal aggregator the other three never use.
ROCStories~\cite{mostafazadeh2016roc,sharma2018tackling} is a standard corpus of
crowd-written five-sentence everyday narratives, each a complete story with a
complication and a resolution, used for commonsense story understanding; the
companion Story Cloze set supplies held-out stories with a labelled correct
ending. We scramble the five sentences and ask the model to recover the true
order. This inverts runtime verification: there one is given the relation and
asked whether a temporal formula holds on it, whereas here the formula is fixed
and the relation is the unknown we learn. Each sentence is
a Kripke world, embedded by a frozen sentence-transformer~\cite{reimers2019sentence};
an \emph{asymmetric} rank-$32$ bilinear head produces a two-place accessibility
$A_\theta(w_i,w_j)=\sigma\!\big(\alpha\,\langle P_{\text{src}}h_i,\,P_{\text{tgt}}h_j\rangle+\beta\big)$
read as ``sentence $s_j$ plausibly follows $s_i$'' ($24{,}578$ trainable parameters
on top of the frozen encoder). For a candidate order $\sigma$ the score is a single
\emph{Until} evaluated at the first world, the adjacent-transition chain
$A_\theta(\sigma(t{-}1),\sigma(t))$ must hold until the last-sentence indicator,
scored over all $5!=120$ permutations of $S_5$ (exact, but factorial in sequence length); training minimises the
contradiction loss directly (as graph colouring does, Section~\ref{sec:exp:coloring}), driving the true (identity) order
above every scramble by a margin $m$.

\vspace{-0.35em}\paragraph{Result and interpretation.} Within a story the five sentences share
characters, vocabulary and tone, so a sentence-transformer cosine ordering sits at
chance ($0.50$/$0.008$); the only signal is ordinal. Over $5$ seeds the bare model recovers
order at pairwise $0.765$ and exact $0.235$ on the held-out
Story-Cloze-2018 set (disjoint by text; in-domain $0.788/0.225$). A stronger frozen
encoder (\textsf{all-mpnet-base-v2}) raises MLNN to 0.819/0.327, on par with a prompted 7B LLM (\texttt{qwen2.5-7b}, 0.854/0.330) and ahead of a pairwise head on the same embeddings (0.809/0.213). The pairwise head learns local order about as well, but not a consistent global order; the \emph{Until} neuron provides that.
Two ablations show what drives the result: removing the contradiction margin lowers pairwise accuracy to 0.708, and tying $P_{\text{src}}=P_{\text{tgt}}$, which makes the kernel symmetric and blind to direction, drops it to 0.566, barely above chance. 
The frame-axiom audit shows what kind of relation this task learns: it is asymmetric (B$=0.87$) and only partly reflexive (T$=0.78$), as a temporal-precedence order should be. This contrasts with C-MAPSS, whose inner-product kernel is symmetric by design; the bilinear head here could have learned a symmetric relation but did not. Transitivity (4$=1.00$) is saturated in both cases and does not distinguish them. % Per-seed ablations, the encoder/LLM comparison and the full audit are in the supplementary material (\emph{Sentence Ordering}) the extended version~\footref{fn:code} shows the axiom validity off four examples directly.

\vspace{-0.35em}\subsection{Graph Colouring: solving from axioms, recovering the constraint graph}\label{sec:exp:coloring}

Graph $k$-colouring is the constraint-satisfaction problem most natural to modal
logic, because the adjacency relation is the accessibility relation:
``adjacent nodes take different colours'' is exactly the modal axiom
$\bigwedge_{c=1}^{k}\big(p_c\to\neg\Diamond p_c\big)$, ``if I am colour $c$,
no accessible node is colour $c$.'' We use graph colouring in both of MLNN's two modes: \emph{Solve}, where $R$ is fixed and $L_{\text{task}}{=}0$ so a colouring is found by minimising the axiom-violation residual, and \emph{Learn}, where the graph is hidden and a learnable $A_\theta$ is recovered from valid colourings.

\vspace{-0.35em}\paragraph{Solve (pure satisfiability mode).} The graph is given, so $R$ is
fixed and there is no task loss ($L_{\text{task}}{=}0$); MLNN simply drives the
per-node axiom-violation residual down until no edge is monochromatic. We test
on planted-$3$-colourable graphs in three tiers ($10$ graphs each; $12$, $20$,
$30$ nodes at rising density), so each rate is out of $10$ per tier and $30$
overall. At that size the intervals are wide and we report them.

MLNN solves $29/30$ ($97\%$, Wilson $95\%$ CI $[0.83,0.99]$). Only one of the
three comparisons is statistically supported. Against the non-modal ablation,
which keeps the identical gradient solve but swaps $\Box/\Diamond$ for a bare
quadratic peer penalty, $18/30$ ($60\%$, $[0.42,0.75]$): the gap is significant (Fisher $p=0.001$), so the modal operator, not the gradient solver, accounts for the gain. Against a Semantic-Loss encoding~\cite{xu2018semantic}, $26/30$ ($87\%$), the difference is \emph{not} significant ($p=0.35$), so we withdraw any ordering between them; and a recurrent relational-net (RRN-style) GNN ties MLNN exactly (29/30, $p=1.0$), which suggests that the gain comes from adding relational structure, not from the modal operator specifically. Exact solvers and metaheuristics solve every instance, and supervised differentiable solvers (SATNet~\cite{wang2019satnet}, RRN~\cite{palm2018rrn}) perform well on the distributions they were trained on. MLNN differs in two ways: it needs no training data, and its per-node residual $L_{\text{contra}}$ shows which nodes still violate the axiom, falling to zero once the colouring is proper. This trajectory and the full baseline comparison are in \emph{Graph Colouring} in the extended version\footref{fn:code}.

\vspace{-0.35em}\paragraph{Learn (inductive accessibility).} 
The inductive variant hides the graph and shows the model only valid colourings. We train a learnable
$A_\theta$ to satisfy the colouring axiom across all of them while pushing
$A_\theta$ as large as the axiom permits; pairs that never share a colour
saturate towards $1$ and pairs that sometimes do are forced towards $0$. From
$60$ proper colourings of a hidden $16$-node graph the learned $A_\theta$
recovers the true adjacency at edge-recovery $AUC=1.000$: the learned
scores rank every true edge above every non-edge, so thresholding recovers the
graph exactly (all $11$ edges, no spurious one) given a sufficiently diverse
colouring set. A fixed-relation solver cannot do this, since its constraint graph is given in advance. Here the accessibility relation that the analyst inspects is learned from data, and it matches the constraint graph exactly.

% This variant is also the paper's one \emph{measurable} test of interpretability: elsewhere a learned relation can only be judged by whether it looks plausible, but here the hidden graph is ground truth, so ``is the artefact the analyst audits correct?'' has a number, and it is exactly correct. Two further measurements, faithfulness and seed-stability ($r=0.84$ across five seeds), are reported in the appendix.
This is also the only experiment in which interpretability can be measured directly. Elsewhere we can only judge whether a learned relation looks plausible; here the hidden graph is known, so we can check if the relation is correct. The extended version\footref{fn:code} also reports faithfulness and seed stability ($r=0.84$ across five seeds).
\vspace{-0.5em}
\section{Discussion and Conclusion}

\label{sec:conclusion}

MLNNs extend differentiable neurosymbolic reasoning to modal logic by making
$(W,R,V)$ a first-class differentiable component with learnable valuation and
accessibility. The same matrix is used for prediction and for inspection. Across the four experiments, we showed how it can be interpreted: a trust matrix, a regime embedding with
deontic bounds, a temporal-precedence order, and a recovered constraint graph.
On task metrics MLNN is competitive but not state of the art, and two results need qualification: the doxastic margin over the LLM-judge baseline is attributable mainly to the NLI front-end, not the modal reasoning; on colouring, a relational GNN performs comparably.
The main contribution is therefore the learned relation itself: it is the relation that produces the answer, and where ground truth exists, it is recovered exactly.

\paragraph{Limitations.} The analyst must choose the worlds and axioms, and a poor choice may be one training cannot repair. Frame axioms are only audited after training, not enforced, and exact top-$k$ masking keeps end-to-end cost at $O(|W|^2)$.

\paragraph{Use of AI Tools:} The idea is entirely human. The first author used AI assistants in this project for proposing and implementing experiments, literature search, drafting, and checking our claims against the outputs.
%
% Furthermore, all future applications of MLNNs are based on real-world industry problems brought forward by the second author, with full adherence to all relevant integrity, privacy, confidentiality, and non-disclosure agreements.

%\clearpage 

% References section - not counted toward 10-page limit
% \bibliographystyle{plain}
% \bibliography{references}
\bibliography{references}

% \clearpage
%\includepdf[pages=-]{supplementary_260905}
\includepdf[pages=-]{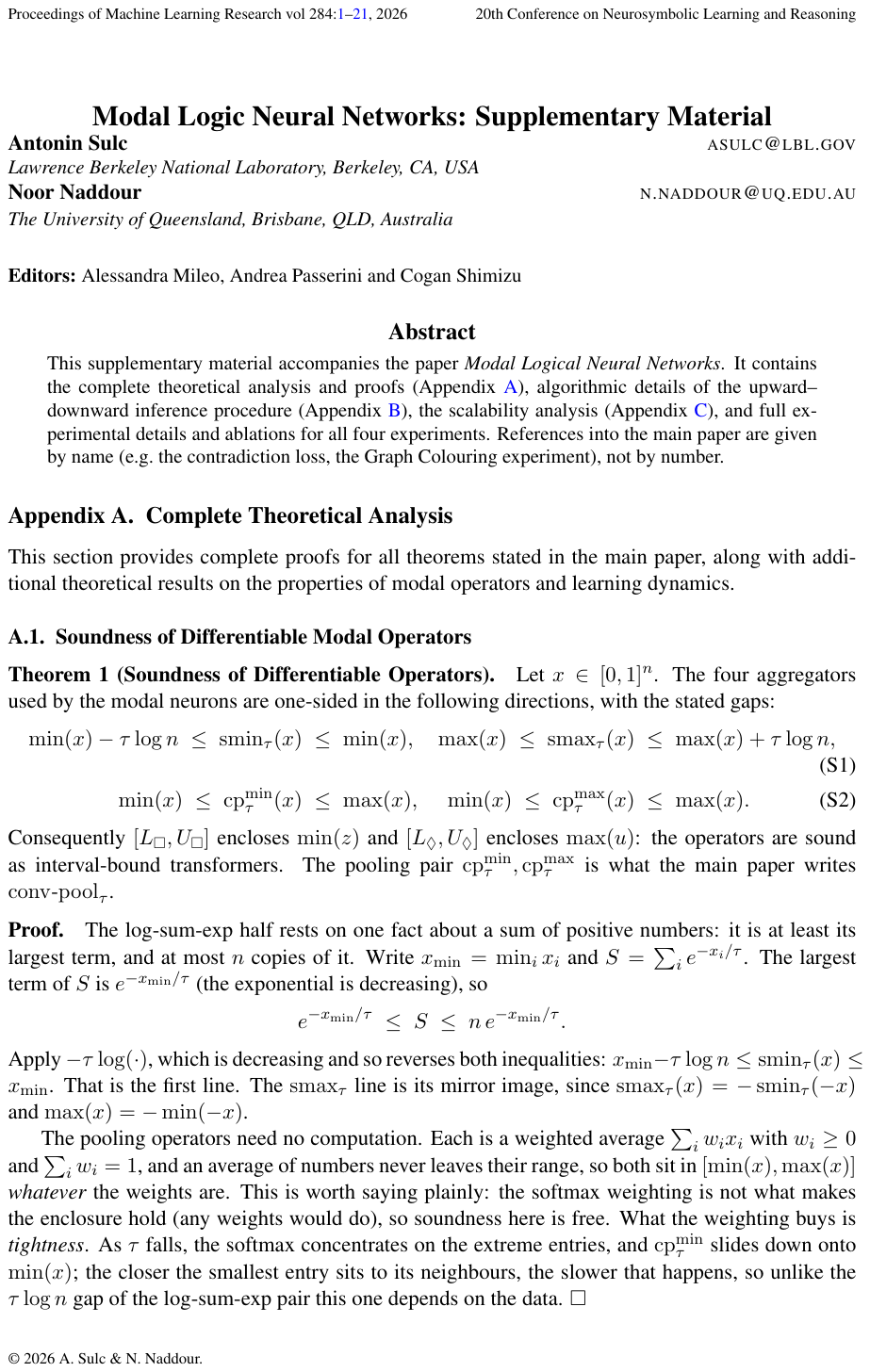}

\end{document}